\documentclass[sn-basic,square,comma,numbers]{sn-jnl}
\setcitestyle{numbers}

\makeatletter
\renewcommand\@biblabel[1]{#1.}
\makeatother
\usepackage{array}
\newcolumntype{P}[1]{>{\centering\arraybackslash}p{#1}}

\usepackage[retain-explicit-plus]{siunitx}
\usepackage{makecell}
\usepackage{lineno}
\usepackage{hyperref}
\usepackage{multirow}
\usepackage{booktabs}
\usepackage{xcolor}

\usepackage{subcaption}
\usepackage{graphicx}

\usepackage{listings}
\usepackage{amssymb}
\usepackage{pifont}
\newcommand{\cmark}{\ding{51}}
\newcommand{\xmark}{\ding{55}}

\jyear{2023}

\modulolinenumbers[5]
\usepackage{blindtext}
\usepackage{hyperref}
\usepackage{nameref}

\newcounter{mylabelcounter}

\makeatletter
\newcommand{\labelText}[2]{%
\refstepcounter{mylabelcounter}%
\immediate\write\@auxout{%
 \string\newlabel{#2}{{\unexpanded{#1}}{\thepage}{{\unexpanded{#1}}}{mylabelcounter.\number\value{mylabelcounter}}{}}%
}%
}
\makeatother

\begin{document}

\title[{Explainable cognitive decline detection in free dialogues}]{Explainable cognitive decline detection in free dialogues with a Machine Learning approach based on pre-trained Large Language Models}

\author*[1]{\fnm{Francisco} \sur{de Arriba-Pérez}}\email{farriba@gti.uvigo.es}

\author[1]{\fnm{Silvia} \sur{García-Méndez}}\email{sgarcia@gti.uvigo.es}

\author[1]{\fnm{Javier} \sur{Otero-Mosquera}}\email{jotero@gti.uvigo.es}

\author[1]{\fnm{Francisco J.} \sur{González-Castaño}}\email{javier@det.uvigo.es}

\affil[1]{\orgname{Information Technologies Group, atlanTTic, University of Vigo}, \city{Vigo}, \country{Spain}}

\abstract{Cognitive and neurological impairments are very common, but only a small proportion of affected individuals are diagnosed and treated, partly because of the high costs associated with frequent screening. Detecting pre-illness stages and analyzing the progression of neurological disorders through effective and efficient intelligent systems can be beneficial for timely diagnosis and early intervention. We propose using Large Language Models to extract features from free dialogues to detect cognitive decline. These features comprise high-level reasoning content-independent features (such as comprehension, decreased awareness, increased distraction, and memory problems). Our solution comprises (\textit{i}) preprocessing, (\textit{ii}) feature engineering via Natural Language Processing techniques and prompt engineering, (\textit{iii}) feature analysis and selection to optimize performance, and (\textit{iv}) classification, supported by automatic explainability. We also explore how to improve Chat\textsc{gpt}'s direct cognitive impairment prediction capabilities using the best features in our models. Evaluation metrics obtained endorse the effectiveness of a mixed approach combining feature extraction with Chat\textsc{gpt} and a specialized Machine Learning model to detect cognitive decline within free-form conversational dialogues with older adults. Ultimately, our work may facilitate the development of an inexpensive, non-invasive, and rapid means of detecting and explaining cognitive decline.}


This version of the article has been accepted for publication, after peer review (when applicable) but is not the Version of Record and does not reflect post-acceptance improvements, or any corrections. The Version of Record is available online at: https://doi.org/10.1007/s10489-024-05808-0.

\maketitle

\section{Introduction}
\label{sec:introduction}

Progressive neurological disorders (\textit{e.g.}, Alzheimer’s disease) affect 40 million people worldwide \citep{Balagopalan2021} and are a common cause of death \citep{AlzheimerAssociation2021,Serkan2022}. However, only \SI{25}{\percent} of affected people receive a diagnosis. There are multiple reasons for this, including stigma and a lack of awareness and resources \citep{Jammeh2018,Mao2023}. The number of older adults with Alzheimer's disease is predicted to rise to 150 million by 2050 \citep{Lynch2020}. Consequently, detecting pre-illness stages and analyzing the progression of neurological disorders using cost-effective and efficient intelligent systems is crucial to ensure timely diagnosis, risk assessment, and early intervention \citep{Serkan2022,Serkan2019}.

Numerous cognitive tests (\textsc{ct}, \textit{e.g.}, Alzheimer’s Disease Assessment Scale-Cognition - \textsc{adasc}og, Cognitive Dementia Rating - \textsc{cdr}, Mini-Mental State Examination - \textsc{mmse}, Montreal Cognitive Assessment - Mo\textsc{ca}, etc.) are currently used to diagnose and monitor neurological disorders, but they need to be applied manually and are therefore costly \citep{Mao2023,Guo2019, Agbavor2022}. These tests can be automated with the help of Artificial Intelligence (\textsc{ai}), enabling more frequent screening of target populations \citep{Graham2020,Alshayeji2023} and the conduct of longitudinal studies. The proposal described in this work consists of an automatic, continuous evaluation of cognitive performance based on engaging dialogues established with end users. Engagement fosters evaluation over time, which is of interest to longitudinal studies. The dialogues are supported by advanced \textsc{ai} techniques.

\textsc{ai}-based solutions for clinical assessment purposes is a fast-growing research field, and numerous studies have already analyzed applications in the fields of progressive neurological disorders and dementia \citep{Salvatore2018,lee2022factors,Merkin2022,Amini2023,sirilertmekasakul2023current}. Machine Learning (\textsc{ml}) models \citep{Balagopalan2021,LauraHernandez2018,Liu2020,Qiao2021} (\textit{e.g.}, Convolutional Neural Networks - \textsc{cnn} \citep{Lin2018}) and Natural Language Processing (\textsc{nlp}) techniques \citep{Orimaye2017,Kepesiova2023,Amini2023,Ying2023} have been applied to textual and voice data. These techniques can effectively predict cognitive impairment and content-dependent and context-independent features can be leveraged to improve the predictive performance of \textsc{ml} models in this area.

Among the multiple bio-markers available for the study of cognitive decline \citep{Molinuevo2018,Villa2020,Zhang2021}, language acquisition is an inexpensive, non-invasive, and readily accessible tool \citep{Agbavor2022}. 
However, language conceals large volumes of information within complex relationships that can be difficult to decipher. Large Language Models (\textsc{llm}s) are better equipped to navigate and process complex information and have therefore received increasing attention in the medical field \citep{Balagopalan2021, lee2020, Roshanzamir2021}. \textsc{llm}s have been used to analyze images and prescribe medical treatments and have also proven capable of passing medical accreditation exams \cite{Hadi2023}. However, personalized medical treatment recommendations provided by \textsc{llm}s remain unreliable \cite{Alberts2023}.

\textsc{llm}s have broader language generation and understanding capabilities \citep{Agbavor2022,Chen2023},  compared to domain-dependent models, when adequate prompt engineering is applied \citep{Chen2024}. One noteworthy recent example of  \textsc{llm} enhanced text generation capabilities is \textsc{gpt-4} \citep{wang2023text}. This \textsc{llm} is used by Chat\textsc{gpt} \citep{Deng2023} to generate coherent dialogues with human users. However, there is still limited research on whether \textsc{llm}s are better than niche solutions based on highly focused, context- or domain-dependent \textsc{ml} models, particularly in the clinical field \citep{wang2023text}. Our approach combines both approaches. It employs an \textsc{llm} to create a friendly conversational assistant environment. The dialogue is augmented by additional services (\textit{e.g.}, weather reports, and medication reminders) to generate interest among the end users and enable a transparent longitudinal evaluation of cognitive decline. This \textsc{llm} is also used to generate high-level reasoning and content-independent side features for a specialized \textsc{ml} model. This combined usage of an \textsc{llm} and \textsc{ml} model for conducting explainable cognitive evaluation is an entirely novel contribution.

In general, establishing trust in \textsc{ai} is necessary to fight the common perception that models are black boxes, leaving end users and developers in the dark about their decision-making processes \citep{Dutt2023}. This issue is especially relevant in medical practice, particularly in personalized treatments \citep{Agbavor2022}. Explainable \textsc{ai} (\textsc{xai}) \citep{BarredoArrieta2020} exploits the intrinsic interpretability of certain \textsc{ml} algorithms (\textit{e.g.}, tree-based models such as Random Forest - \textsc{rf}), or implements methods to bypass the opacity of non-interpretable models \citep{akbar2023trustworthy}. \textsc{xai} techniques include feature importance \citep{FranciscoDeArriba2022}, counterfactual explanations \citep{Wachter2017}, natural language descriptions \citep{Ehsan2019}, and visual representations \citep{Spinner2019}. Explainability in \textsc{llm} models is far from being solved. Our proposal combines \textsc{llm}s with explainable \textsc{ml} classifiers to generate an explanation of the predictions that caregivers or healthcare experts will be able to understand.

Summing up, the main contribution of this work is a hybrid \textsc{ml} solution for assessing cognitive state, which combines context-dependent features with an \textsc{llm} for generating high-level reasoning, context-independent features. The latter provides high-level reasoning on the behavior of a user who dialogues with a conversational assistant on a leisure topic. Moreover, these features greatly enhance the explainability of the system decisions about cognitive state compared to rigid content-dependent features.

The rest of this paper is organized as follows. Section \ref{sec:related_work} reviews the relevant competing works on cognitive decline detection involving \textsc{llm}s, and Section \ref{sec:contributions} summarizes the contribution of this work beyond state of the art. Section \ref{sec:methodology} explains the proposed solution. Section \ref{sec:results} describes the experimental data set, our implementations, and the results obtained. Finally, Section \ref{sec:conclusions} concludes the paper and proposes future research.

\section{Related work}
\label{sec:related_work}

\textsc{ai} solutions for healthcare seek to enhance diagnosis and treatment while simultaneously optimizing resources \citep{Yu2018}. One promising line in this regard is the development of intelligent assistants or chatbots \citep{Kim2023,padovan2023chatgpt}. A representative example is the work by \citet{Kurtz2023}. They presented a cognitive decline detection solution based on a voice assistant. The authors exploited lexical and semantic features, and embeddings for that purpose.

Although many \textsc{nlp}-based speech analysis solutions exist for the early prediction of Alzheimer's disease \citep{Eyigoz2020,Voleti2020}, there is limited, preliminary research on the use of \textsc{llm}s in the field, apart from specific use cases \citep{Agbavor2022} and medical applications \citep{hristidis2023chatgpt}.

\textsc{llm}s have the ability to offer powerful \textsc{nlp} functionalities that can be utilized in various medical tasks \citep{jethani2023evaluating}. They have demonstrated clinical reasoning abilities \citep{Lee2023}, passed medical licensing exams \citep{wang2023performance}, provided medical advice \citep{Ayers2023}, and even generated clinical notes \citep{Cascella2023}.

The following prior works explore using \textsc{llm}s for cognitive evaluation.

\citet{Yuan2020} applied \textsc{bert} and \textsc{ernie} models to model disfluencies and language problems in patients with Alzheimer's disease. The experimental data consisted of transcriptions from cognitive tests in the \textsc{adr}e\textsc{ss} (Alzheimer’s Dementia Recognition through Spontaneous Speech) data set. However, after fine-tuning, they relied mainly on word embeddings extracted from the \textsc{llm}s. The only side (content-independent) features considered were word frequency and speech pauses, thereby limiting the generalizability of the results when non-standard tests were used. \citet{Qiao2021} extracted disfluency measures and combined them into a stacking classification approach. The absence of content- and context-independent side features makes this approach less robust than ours in settings prone to contextual changes.

\citet{Roshanzamir2021} combined transformer-based deep neural network language models (\textsc{bert}, \textsc{xlm}, \textsc{xln}et) with \textsc{ml} (Logistic Regression - \textsc{lr}, Long Short-Term Memory - \textsc{lstm}, \textsc{cnn}) to detect Alzheimer’s disease using image description testing. Their approach exclusively used embeddings extracted from transcripts and content-dependent features. \citet{zhu2021wavbert} screened for dementia using both non-semantic (speech pauses) and semantic features (word embeddings) extracted from cognitive test speech data by \textsc{bert}. As in the work by \citet{Yuan2020}, the solution was fine-tuned with non-semantic information. However, compared to our solution, the approaches in both \cite{Roshanzamir2021} and \cite{zhu2021wavbert} were highly dependent on semantic information.

\citet{Li2023} estimated the ability of Llama and Chat\textsc{gpt} \textsc{llm}s to detect cognitive impairment from electronic health record (\textsc{ehr}) notes. The prediction outcome, combined with manual assessments by experts, was used to fine-tune the \textsc{bert} model. However, no information is provided about the system's performance without expert backup.

\citet{Agbavor2022} predicted dementia from standard cognitive tests driven by \textsc{gpt-3}. The authors both classified and predicted the severity of Alzheimer's disease. For the classification task, they employed traditional \textsc{ml} classifiers (\textit{i.e.}, \textsc{lr}, Support Vector Classifier - \textsc{svc}, and \textsc{rf}) with acoustic and word embedding features. Regression analysis was then performed using the same features to predict \textsc{mmse} scores. Therefore, the training process was based on user interactions in standard tests. \citet{Mao2023} followed a similar approach to that used in \cite{Roshanzamir2021} but used a Linear Classifier (\textsc{lc}). These studies were based on clinical tests or doctors' notes, limiting their generalizability to non-clinical data. As in previous research, they were strongly dependent on context-dependent word embeddings. Conversely, our proposal uses content-independent side features to focus on longitudinal dialogues on all topics of interest.

Instead of word embeddings extracted with \textsc{llm}s, we propose using high-level reasoning features generated in response to questions presented to the \textsc{llm}s, inspired by human reasoning and independent of particular conversations (\textit{e.g.}, the language register of the user, \textit{i.e.}, adult, child, elder, etc.). \citet{wang2023text} followed a similar approach. However, they tasked Chat\textsc{gpt} to extract high-level features from the DementiaBank data set instead of using free dialogues, as in our case. None of the previous works discussed provided explainability capabilities.

\subsection{Contributions}
\label{sec:contributions}

We propose using the Chat\textsc{gpt} \textsc{llm} to extract high-level reasoning, content-independent side features from loosely driven entertainment dialogues with a chatbot (as explained in Section \ref{sec:dataset}). We identified relevant features by analyzing previous research on cognitive function decline (\textit{e.g.}, changes in the content, comprehension, decreased awareness, increased distraction, memory problems, etc.) \citep{Balagopalan2021,Serkan2022,Guo2019,Roshanzamir2021,Mueller2018}. In addition to being context-independent and thus adequate for free dialogue, unlike word-embedding or content-dependent features, the side features used in our proposal support much more interpretable descriptions of the decisions on cognitive decline. As shown in Table \ref{tab:comparison}, few studies have applied \textsc{xai} techniques to our target problem. Although \citet{Bellantuono2022} did not use \textsc{llm}s, they combined an \textsc{rf} classification model for dementia with SHapley Additive ExPlanations (\textsc{shap}) techniques to infer the contributions of the features to the model's predictive performance. A similar approach was applied by \citet{Lombardi2022}, who stated that \textsc{xai} was still in its infancy in computational neuroscience.

Summing up, our work is the first to apply \textsc{ml} techniques to high-level reasoning features extracted from free-form dialogues using an \textsc{llm} to detect cognitive decline. The \textsc{ml} techniques used were Naive Bayes - \textsc{nb}, Decision Tree - \textsc{dt}, and \textsc{rf}, and we further describe the decisions using explainability techniques. 

\begin{table}[!htbp]
\centering
\caption{Comparison of related solutions taking into account the \textsc{llm} and \textsc{ml} models used (\textsc{n/a}: not applicable), the context of the input data (\textsc{ct}: cognitive tests, \textsc{ehr}s: electronic health records), the features involved (\textsc{hlr}: high-level reasoning), and explainability capability (Exp.). ``Side'' refers to context-independent features and content to context-dependent features.}
\label{tab:comparison}
\begin{tabular}{cccccc}
\toprule
\textbf{Authorship} & \textbf{LLM} & \textbf{ML} & \textbf{Input} & \textbf{Features} & \textbf{Exp.}\\
\midrule

\citet{Mao2023} & \textsc{bert}& \textsc{lc} & \textsc{ehr}s & Content & \xmark\\\midrule

\multirow{3}{*}{\citet{Agbavor2022}} & \multirow{3}{*}{\textsc{gpt-3}}& \multirow{3}{*}{\makecell{\textsc{lr} \\ \textsc{svc} \\ \textsc{rf}}} & \multirow{3}{*}{\textsc{ct}} & \multirow{3}{*}{\makecell{Side \\ Content}} & \multirow{2}{*}{\xmark}\\\\\\\midrule

\multirow{2}{*}{\citet{Qiao2021}} & \multirow{2}{*}{\textsc{bert}}& \multirow{2}{*}{\makecell{\textsc{lr} \\ \textsc{cnn}}} & \multirow{2}{*}{\textsc{ct}} & \multirow{3}{*}{\makecell{Side \\ Content}} & \multirow{2}{*}{\xmark}\\\\\\\midrule

\multirow{3}{*}{\citet{Roshanzamir2021}} & \multirow{3}{*}{\makecell{\textsc{bert} \\ \textsc{xlm} \\ \textsc{xln}et}} & \multirow{3}{*}{\makecell{\textsc{lr} \\ \textsc{lstm} \\ \textsc{cnn}}} & \multirow{3}{*}{\textsc{ct}} & \multirow{3}{*}{Content} & \multirow{3}{*}{\xmark}\\\\\\\midrule

\citet{wang2023text} & Chat\textsc{gpt}& \textsc{n/a} & \textsc{ct} & Side & \xmark\\\midrule

\multirow{2}{*}{\citet{Yuan2020}} & \multirow{2}{*}{\makecell{\textsc{bert} \\ \textsc{ernie}}} & \multirow{2}{*}{\textsc{n/a}}& \multirow{2}{*}{\textsc{ct}} & \multirow{2}{*}{\makecell{Side \\ Content}} & \multirow{2}{*}{\xmark}\\\\\midrule

\multirow{2}{*}{\citet{zhu2021wavbert}} & \multirow{2}{*}{\textsc{bert}}& \multirow{2}{*}{\textsc{n/a}} & \multirow{2}{*}{\textsc{ct}} & \multirow{2}{*}{\makecell{Side \\ Content}} & \multirow{2}{*}{\xmark}\\\\\midrule

\multirow{3}{*}{\citet{Li2023}} & \multirow{3}{*}{\makecell{\textsc{llama} \\ Chat\textsc{gpt} \\ \textsc{bert}}} & \multirow{3}{*}{\textsc{n/a}} & \multirow{3}{*}{\textsc{ehr}s} & \multirow{3}{*}{Content} & \multirow{3}{*}{\xmark}\\\\\\\midrule

\multirow{3}{*}{\textbf{Our proposal}} & \multirow{3}{*}{\makecell{Chat\textsc{gpt}}}& \multirow{3}{*}{\makecell{\textsc{nb} \\ \textsc{dt} \\ \textsc{rf}}} & \multirow{3}{*}{\makecell{Free \\ dialogues}} & \multirow{3}{*}{\makecell{Side \\ (linguistic+\textsc{hlr})}} & \multirow{3}{*}{\cmark}\\\\\\

\bottomrule
\end{tabular}
\end{table}

\section{Methodology}
\label{sec:methodology}

\begin{figure}[!htpb]
 \centering
 \includegraphics[scale=0.133]{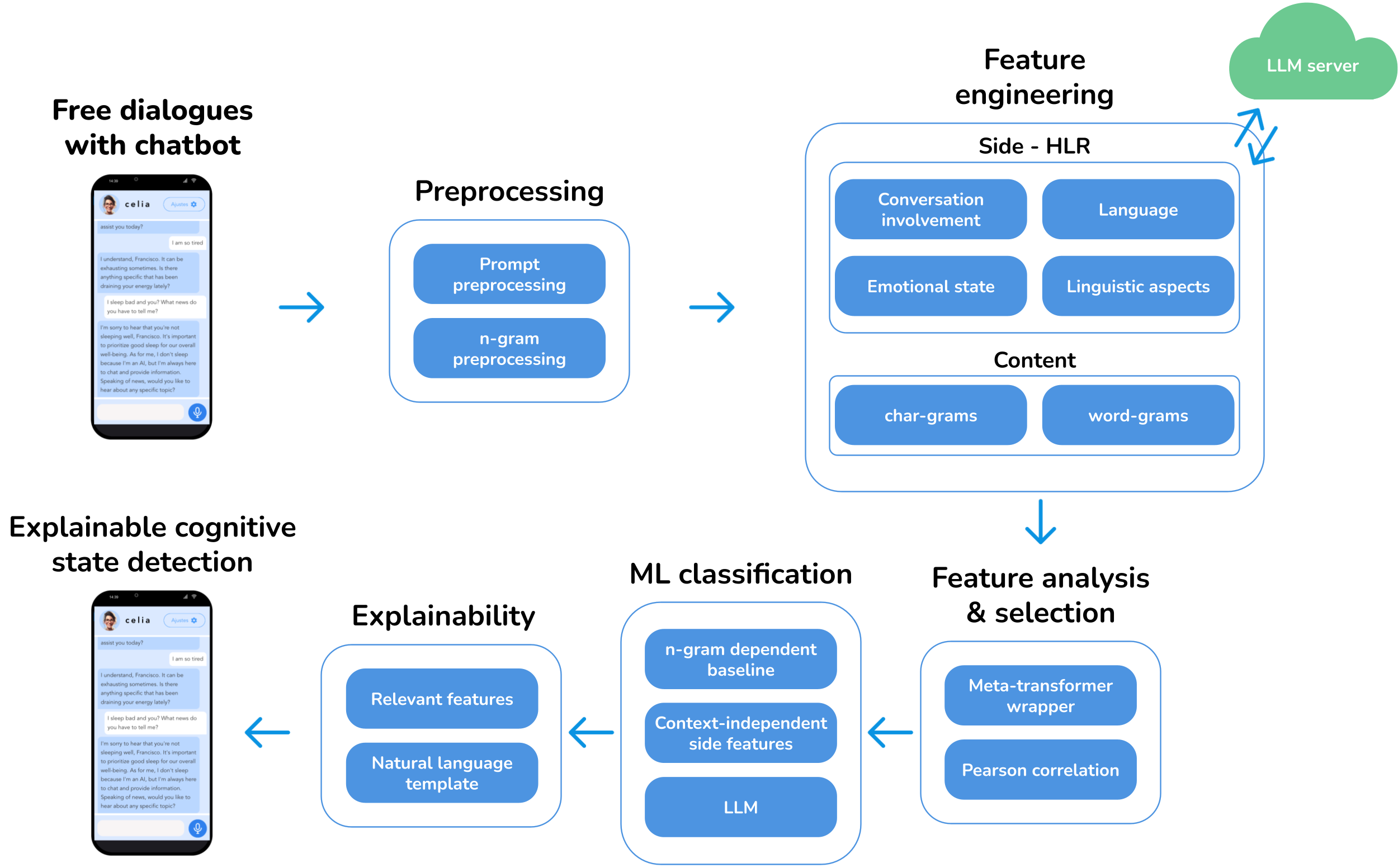}
 \caption{System scheme.}
 \label{fig:scheme}
\end{figure}

Figure \ref{fig:scheme} illustrates the proposed solution for assessing cognitive impairment using free dialogues with older adults. The solution is composed of (\textit{i}) a preprocessing module to prepare the content for further analysis, (\textit{ii}) a feature engineering module that generates an appropriate and comprehensive set of features using \textsc{nlp} techniques to train the cognitive design classification models, (\textit{iii}) a feature analysis and selection module, and (\textit{iv}) a classification module, which is evaluated using standard \textsc{ml} metrics (accuracy, precision, recall) and provides explainability results. Finally, the most representative features of the \textsc{ml} model were selected to evaluate if they can enhance the direct cognitive impairment prediction capabilities of the Chat\textsc{gpt} \textsc{llm} by means of prompt engineering.

To ensure reproducibility, the specific methods used to implement all the steps mentioned and their configuration parameters are specified in Section \ref{sec:results}.

\subsection{Preprocessing module}
\label{sec:preprocessing}

The conversations used in this research are free dialogues between users and an intelligent conversational assistant \citep{deArriba-Perez2022}, organized into daily sessions and automatically transcribed into text. The preprocessing module is essential for ensuring the quality of the input data in the feature engineering process involving both prompt engineering and $n$-gram generation. For prompt engineering, \textsc{nlp} techniques are applied to remove emoticons and hashtags. For $n$-gram generation, images, links, and special characters (\textit{e.g.}, \&, \$, \texteuro) are removed using regular expressions. Then, the textual content is tokenized and lemmatized. Because this use case is based on free dialogues, an \textit{ad hoc} stopword list was created to exclude the terms `no', `yes' (\textit{si}), `more' (\textit{m\'as}), `but' (\textit{pero}), `very' (\textit{muy}), `without'(\textit{sin}), `much' (\textit{mucho}), `little', (\textit{poco}) and `nothing' (\textit{nada}). The final step is the removal of numeric values, isolated characters, and accents.

\subsection{Feature engineering module}
\label{sec:feature_engineering}

Once the textual content of the user's utterances is processed, the feature engineering module generates a broad set of features, detailed in Table \ref{tab:features}. Each entry in the data set corresponds to a user dialogue session. Special attention is paid to user engagement (features 1-8), emotional state (features 9-12), language used (features 13-22), and other linguistic aspects (features 23-26), such as the use of polar responses and bad/complex words. The measurements that produce features 1-24 are within the value range $(0,1)$, where 0 is the minimum and 1 the maximum (except for feature 11, for which the measurements take values $\{0,0.5,1\}$\footnote{Respectively corresponding to the typical sentiment categories $\{-1,0,1\}$ to equalize the ranges of all the measurements.}). Features 25-26 are directly based on counters. Features 1-26 are calculated using an \textsc{llm} and prompt engineering. The measurements of features 1-24 are first computed for each machine-user utterance pair in the session dialogue. Then, each final feature is a list comprising a) the mean, maximum, minimum, and three quartile values (\SI{25}{\percent}, median, \SI{75}{\percent}) of the corresponding measurements across all the machine-user utterance pairs of the current user session plus the whole dialogue in the session and b) the mean, maximum, minimum, and three quartile values of these same statistics based on the current session and all past sessions of the same user (session history). Features 25-26 are first calculated as lists of the mean, maximum, minimum, and three quartile values only for the user utterances in the session (that is, ignoring the machine utterances) and then computed again with the mean, maximum, minimum, and three quartile values of these same statistics based on the session history of the same user. In other words, features 1-24 are lists of forty-nine real values (which we call {\it components} of those features), while features 25-26 are lists of forty-two real values.

Content-independent features 1-26 can be used to assess cognitive deterioration based on free dialogues with the conversational assistant. Linguistic aspects (features 23-26 in Table \ref{tab:features}) and characteristics of the language used (features 13-22) are independent of the conversation topic but allow us to measure the quality and coherence of the user responses. Exclusively for comparison purposes, these features are supplemented with content-dependent features based on char-grams and word-grams extracted from all user utterances from the current session (features 27-28). Algorithm \ref{alg:feature_eng} describes the complete feature engineering process.

\begin{algorithm*}[!htbp]
 \caption{\label{alg:feature_eng}: {\bf Feature engineering}}
 \begin{algorithmic}[0]
 \Function{features\_1\_26}{session, session\_history\_matrix}
     
     \State session=preprocessing\_text(session)
     \State features\_matrix=[[ ]]
     \For{utterance in session}
       \State features\_1\_24 = LLM\_prompt(utterance[``bot"],utterance[``user"])
        \State features\_matrix.add\_row(features\_1\_24)
     \EndFor

    \State features\_vector=[ ]
    \For{column in features\_matrix}
        \State [mean, maximum, minimum, q1, q2, q3] = feature\_statistics(column)
        \State features\_vector.expand([mean, maximum, minimum, q1, q2, q3])
    \EndFor
    \State features\_1\_24\_whole\_dialogue = LLM\_prompt(session[``whole\_dialogue"])
    \State features\_vector.expand(features\_1\_24\_whole\_dialogue)

    \State features\_matrix=[[ ]]
     \For{utterance in session}
       \State features\_25\_26 = LLM\_prompt(utterance[``user"])
        \State features\_matrix.add\_row(features\_25\_26)
     \EndFor
    \For{column in features\_matrix}
        \State [mean, maximum, minimum, q1, q2, q3] = feature\_statistics(column)
        \State features\_vector.expand([mean, maximum, minimum, q1, q2, q3])
    \EndFor
    \State session\_history\_matrix.add\_row(features\_vector)
    
    \State component\_vector=[]
    \For{column in session\_history\_matrix}
        \State [mean, maximum, minimum, q1, q2, q3] = feature\_statistics(column)
        \State component\_vector.expand([mean, maximum, minimum, q1, q2, q3])
    \EndFor
    \State \textbf{return} component\_vector
 
 \EndFunction

 \Function{features\_27\_28}{session}

    \State user\_text=[[ ]]
    \For{utterance in session}
        \State text\_preprocessed = preprocessing\_text(utterance[``user"])
        \State user\_text.add\_row(text\_preprocessed)
    \EndFor
    \State n\_grams = n\_grams\_generation(user\_text)
    \State \textbf{return} n\_grams

 \EndFunction
 
 \end{algorithmic}
\end{algorithm*} 

\begin{table}[!htbp]
\centering
\footnotesize
\caption{\label{tab:features}Features engineered to detect cognitive decline in free dialogues.}
\begin{tabular}{llm{2cm}lc}
\toprule
\bf Type & \bf ID & \bf Name & \bf Description & \bf Source\\\hline

\multirow{46}{*}{Side} & 1 & Amnesia & \begin{tabular}[c]{@{}p{4.2cm}@{}} Inability to recall past events, experiences or information.\end{tabular}& \multirow{42}{*}{\begin{tabular}[c]{@{}p{2.0cm}@{}} \centering Machine-user utterance pairs / whole dialogue \end{tabular}}\\

& 2 & Disconnection & \begin{tabular}[c]{@{}p{4.2cm}@{}} Lack of engagement.\end{tabular}&\\

& 3 & Fatigue & \begin{tabular}[c]{@{}p{4.2cm}@{}} Sense of exhaustion and tiredness.\end{tabular}&\\

& 4 & Hesitation & \begin{tabular}[c]{@{}p{4.2cm}@{}} Uncertainty in responding to a question.\end{tabular}&\\

& 5 & Initiative & \begin{tabular}[c]{@{}p{4.2cm}@{}} Active participation in the conversation.\end{tabular}&\\

& 6 & Naturalness & \begin{tabular}[c]{@{}p{4.2cm}@{}} Smooth and 
effortless flow of communication.\end{tabular}&\\

& 7 & Relax & \begin{tabular}[c]{@{}p{4.2cm}@{}} Sense of ease and comfort.\end{tabular}&\\

& 8 & Stress & \begin{tabular}[c]{@{}p{4.2cm}@{}} Sense of mental or emotional strain.\end{tabular}&\\

\cmidrule{2-4}

& 9 & Happiness & \begin{tabular}[c]{@{}p{4.2cm}@{}} Sense of joy.\end{tabular}&\\

& 10 & Mood swings & \begin{tabular}[c]{@{}p{4.2cm}@{}} Changes in emotional state or mood.\end{tabular}&\\

& 11 & Polarity & \begin{tabular}[c]{@{}p{4.2cm}@{}} Sentiment of the content (negative, neutral, positive).\end{tabular}&\\

& 12 & Sadness & \begin{tabular}[c]{@{}p{4.2cm}@{}} Sense of sorrow.\end{tabular}&\\

\cmidrule{2-4}

& 13 & Adult register & \begin{tabular}[c]{@{}p{4.2cm}@{}} Use of mature language and clear expression of ideas.\end{tabular}&\\

& 14 & Childlike register & \begin{tabular}[c]{@{}p{4.2cm}@{}} Use of simple language and expression of curiosity.\end{tabular}&\\

& 15 & Colloquial register & \begin{tabular}[c]{@{}p{4.2cm}@{}} Use of informal language and incomplete sentences.\end{tabular}&\\

& 16 & Comprehension problems & \begin{tabular}[c]{@{}p{4.2cm}@{}} Difficulty understanding information.\end{tabular}&\\

& 17 & Elder register & \begin{tabular}[c]{@{}p{4.2cm}@{}} Relation of past events and life experiences.\end{tabular}&\\

& 18 & Expression problems & \begin{tabular}[c]{@{}p{4.2cm}@{}} Difficulty in expressing ideas.\end{tabular}&\\

& 19 & Formal register & \begin{tabular}[c]{@{}p{4.2cm}@{}} Use of polite language and complete sentences.\end{tabular}&\\

& 20 & Repetitive language & \begin{tabular}[c]{@{}p{4.2cm}@{}} Lack of a clear flow in the conversation.\end{tabular}&\\

& 21 & Rush language & \begin{tabular}[c]{@{}p{4.2cm}@{}} Impatience or urgency in responses.\end{tabular}&\\

& 22 & Short response & \begin{tabular}[c]{@{}p{4.2cm}@{}} Concise and brief responses.\end{tabular}&\\

\cmidrule{2-4}

& 23 & Bad words & \begin{tabular}[c]{@{}p{4.2cm}@{}} Use of bad language to show strong emotions.\end{tabular}&\\

& 24 & Complex words & \begin{tabular}[c]{@{}p{4.2cm}@{}} Use of a higher level of vocabulary and knowledge.\end{tabular}&\\

\cmidrule{2-5}

& 25 & Polar response counters & \begin{tabular}[c]{@{}p{4.2cm}@{}} Number of polar answers (yes or no) given by the user in the textual content.\end{tabular}& \multirow{4}{*}{\begin{tabular}[c]{@{}p{2.0cm}@{}}User utterances \end{tabular}}\\

& 26 & \textsc{pos} counters & \begin{tabular}[c]{@{}p{4.2cm}@{}} Number of adjectives, adverbs (negative and positive), nouns, and verbs in the textual content.\end{tabular}&\\

\cmidrule{1-5}

\multirow{3}{*}{Content} & 27 & Char-grams & \begin{tabular}[c]{@{}p{4.2cm}@{}} List of char $n$-grams in the content.\end{tabular}& \multirow{2}{*}{\begin{tabular}[c]{@{}p{2.0cm}@{}}Whole dialogue\end{tabular}}\\

& 28 & Word-grams & \begin{tabular}[c]{@{}p{4.2cm}@{}} List of word $n$-grams in the content.\end{tabular}&\\

\bottomrule
\end{tabular}
\end{table}

\subsection{Feature analysis \& selection module}
\label{sec:feature_selection}

In the feature analysis and selection stage, irrelevant and other features that could degrade the system's performance are removed to ensure optimal training of the \textsc{ml} classifiers. We have implemented two feature analysis and selection techniques: (\textit{i}) a relevance-based technique using a tree algorithm and (\textit{ii}) correlation analysis with the Pearson coefficient.

\begin{description}

\item A meta-transformer wrapper based on a tree-based \textsc{ml} model is applied to select the most significant feature components for training the classification module, regardless of the specific \textsc{ml} model used (thus, we follow a model-agnostic strategy \citep{Burkart2021}). The wrapper, using the Mean Decrease in Impurity (\textsc{mdi}) technique, leverages importance weights to identify and eliminate features based on their impurity contributions \citep{Breiman2017}. This technique calculates the average proportion of sample splits of each feature in each node across all the trees within the tree-based \textsc{ml} model used. Subsequently, feature components with an \textsc{mdi} lower than the average \textsc{mdi} are excluded from further consideration in the classification module.

\item To select the final set of feature components for the \textsc{llm} prompt engineering stage (scenario 3 in the next section), we calculated the Pearson correlation coefficient \citep{Benesty2009}, where -1 and 1, respectively, represented the two variables with the strongest possible inverse and direct relationship between the most relevant feature components as previously described and the target (cognitive decline). We then selected the components that exceeded a correlation threshold.

\end{description}

\subsection{Classification module}
\label{sec:evaluation}

Three different scenarios are considered:

\begin{description}
 \item \textbf{Scenario 1} trains an \textsc{ml} classification based exclusively on content-dependent textual features derived from user utterances (features 27-28 in Table \ref{tab:features}). This scenario is intended to evaluate $n$-grams as a source, as in prior works, \cite{Qiao2021,Orimaye2017,Kepesiova2023} and serves as a baseline for our study.
 
 \item \textbf{Scenario 2} trains an \textsc{ml} classification considering context-independent side features 1-26 as defined in Table \ref{tab:features}, where the base measurements are calculated by applying an \textsc{llm} to user utterances, the machine-user utterance pairs and the whole dialogue, depending on the case.
 
 \item \textbf{Scenario 3} evaluates the performance of the \textsc{llm} as a classifier of cognitive impairment \cite{Li2023,wang2023text} using two approaches: (\textit{a}) the \textsc{llm} is directly used to analyze the dialogue of a user session as a whole and guess the target ``cognitive decline'' with prompt engineering, and (\textit{b}) the same setting plus prompt enhancements based on a pick of the most relevant features from scenario 2, using the Pearson coefficient as described in the previous section.

\end{description}

In summary, scenario 1 is our baseline, and scenario 2 is used to evaluate our approach. Scenario 3 uses an \textsc{llm} exclusively to directly classify cognitive impairment and compare its performance with our proposal, in which the \textsc{llm} is employed to generate high-level reasoning features.

In scenarios 1 and 2, we use the meta-transformer wrapper to identify the most significant features (see Section \ref{sec:feature_selection}). 

In scenarios 2 and 3, we apply prompt engineering. The prompts are divided into a context section, a request, and the output format. Specific prompts are used in scenario 2 to generate the measurements for the context-independent side features from the textual input (both for features 1-24 and counter-type features 25-26). In scenario 3, two prompts directly address the cognitive impairment level. The first prompt directly applies to the whole dialogue to assess cognitive decline. The second adds a pick of the most relevant features from scenario 2, as previously explained, to the context section of the first prompt. Algorithms \ref{alg:scenario1}, \ref{alg:scenario2}, \ref{alg:scenario3a} and \ref{alg:scenario3b} respectively describe the step-by-step processes in scenarios 1, 2, 3a and 3b.

The algorithms \textsc{nb}, \textsc{dt}, and \textsc{rf} \textsc{ml} were selected based on their favorable performance in related studies in the literature \citep{Xu2018,Kadhim2019}.

\begin{algorithm*}[!htbp]
 \caption{\label{alg:scenario1}: {\bf Scenario 1 pseudocode}}
 \begin{algorithmic}[0]
 \Function{cognitive\_impairment\_detection\_scenario\_1}{session, classifier}
 
    \State n\_grams = features\_27\_28(session)
    \State selected\_features = meta-transformer\_wrapper\_selector(n\_grams)
    \State prediction = classifier.predict(selected\_features)
    \State \textbf{return} prediction
    
 \EndFunction
 \end{algorithmic}
\end{algorithm*} 

\begin{algorithm*}[!htbp]
 \caption{\label{alg:scenario2}: {\bf Scenario 2 pseudocode}}
 \begin{algorithmic}[0]
 \Function{cognitive\_impairment\_detection\_scenario\_2}{session, session\_history\_matrix, classifier}
 
    \State component\_vector = features\_1\_26(session, session\_history\_matrix)
    \State selected\_features = meta-transformer\_wrapper\_selector(component\_vector)
    \State prediction = classifier.predict(selected\_features)
    \State \textbf{return} prediction, selected\_features
    
 \EndFunction
 \end{algorithmic}
\end{algorithm*} 

\begin{algorithm*}[!htbp]
 \caption{\label{alg:scenario3a}: {\bf Scenario 3a pseudocode}}
 \begin{algorithmic}[0]
 \Function{cognitive\_impairment\_detection\_scenario\_3a}{session}
   
    \State session=preprocessing\_text(session)
    \State prediction = LLM\_prompt\_predict(session[``whole\_dialogue"])
    \State \textbf{return} prediction

 \EndFunction
 \end{algorithmic}
\end{algorithm*} 

\begin{algorithm*}[!htbp]
 \caption{\label{alg:scenario3b}: {\bf Scenario 3b pseudocode}}
 \begin{algorithmic}[0]
 \Function{cognitive\_impairment\_detection\_scenario\_3b}{session, session\_history\_matrix, classifier}
    
    \State session=preprocessing\_text(session)
    \State selected\_features\_scenario\_2 = cognitive\_impairment\_detection\_scenario\_2 (session, session\_history\_matrix, classifier)[selected\_features]
    
    \State selected\_features = pearson\_correlation(selected\_features\_scenario\_2)
    \State prediction = LLM\_prompt\_predict(session[``whole\_dialogue"], selected\_features)
    \State \textbf{return} prediction

 \EndFunction
 \end{algorithmic}
\end{algorithm*} 

\subsection{Explainability module}
\label{sec:explainability}

Natural language explanations about decisions on users' cognitive state are based on the relevant feature components in scenario 2 (the components selected by the meta-transformer wrapper). The relevant features are arranged in descending order of relevance. Note that counter-type features 25-26 are normalized by the number of words. For each decision to be explained, six components of features 1-26 with the highest and lowest values, three of each type, are employed. In the case of a tie components are chosen randomly for the explanation template.

\section{Evaluation and discussion}
\label{sec:results}

This section provides an overview of the experimental data set, describes the implementations to facilitate their reproducibility, and presents the results achieved. The experiments were conducted on a computer with the following specifications.

\begin{itemize}
 \item \textbf{Operating System}: Ubuntu 22.10 \textsc{lts} 64 bits
 \item \textbf{Processor}: \textsc{intel}\textsuperscript\textregistered\ Core\textsuperscript\texttrademark\ i7-12700K
 \item \textbf{RAM}: 32 \textsc{gb} \textsc{ddr4} 
 \item \textbf{Disk}: 1 \textsc{tb} \textsc{ssd}
\end{itemize}

\subsection{Experimental data set}
\label{sec:dataset}

\begin{figure}[!htpb]
 \centering
 \includegraphics[scale=0.20]{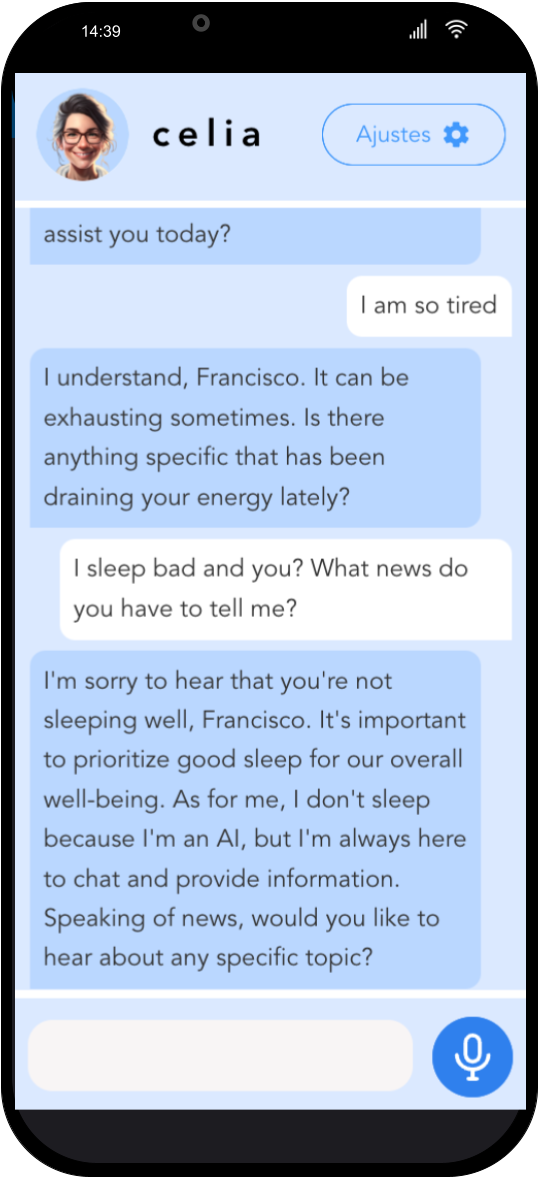}
 \caption{Celia web application (English example).}
 \label{fig:celia}
\end{figure}

The dialogues were collected via the Celia web application\footnote{Available at \url{https://celiatecuida.com}, February 2024.} (see Figure \ref{fig:celia}), and transcribed to text using the Google Cloud Speech-to-Text library\footnote{Available at \url{https://cloud.google.com/speech-to-text}, February 2024.}. Celia has been designed to entertain and accompany elderly people. A rich dialogue is achieved by combining questions generated through an \textsc{llm} with services such as weather, curiosities, and saint days, which vary throughout the year. This allows the free generation of a dynamic conversation on varying topics based on user preferences.

The experimental data set\footnote{Data are available on request from the authors. The data set also includes information on visual, movement, and hearing impairments and information about medication intake.} consists of \num{8220} utterances in Spanish from \num{523} sessions held with \num{42} users registered in the application. Of these, \num{14} had cognitive impairment. Each user participated in an average of \num{12.45} sessions, with a standard deviation of $\pm$ \num{6.32} sessions, and each session had on average \num{15.72} utterances, with a standard deviation of $\pm$ \num{4.89} utterances. Each session had 67.95 words on average, with a standard deviation of $\pm$ 70.14 words. Utterances by people with cognitive impairment comprised 49.55 words $\pm$ 43.74 on average. This represents a reduction of \SI{38.41}{\percent} compared with people without this condition (80.45 words $\pm$ 81.15). Table \ref{tab:dataset_distribution} details the percentages of sessions held with users with and without cognitive impairment. As seen, the data set was rather balanced.

\begin{table}[!htpb]\centering
\caption{Experimental data set distribution.}
\label{tab:dataset_distribution}
\begin{tabular}{lc}\toprule
\textbf{Cognitive impairment} & \multicolumn{1}{l}{\textbf{\% of sessions}} \\
\midrule
Absent & 56.98 \\
Present & 43.02 \\ 
\bottomrule
\end{tabular}
\end{table}

\subsection{Preprocessing module}
\label{sec:preprocessing_results}

Hashtags, images, links, and special characters were identified using the regular expressions in Listing \ref{lst:regular_expressions} and removed. Emoticons and accents were deleted using the \textsc{nfkd} Unicode normalization form\footnote{Available at \url{https://unicode.org/reports/tr15}, February 2024.}. The \textsc{nltk} stop word list was used to delete stop words\footnote{Available at \url{https://www.nltk.org}, February 2024.}. Finally, textual content was lemmatized using the {\tt es\_core\_news\_md} core model from the \texttt{spaCy} Python library\footnote{Available at \url{https://spacy.io/models/es}, February 2024.}.

\begin{lstlisting}[frame=single,caption={Regular expression applied to detect hashtags, images, links, and special characters.}, label={lst:regular_expressions},emphstyle=\textbf,escapechar=ä]
ä\textbf{Hashtags}ä = rä`\textbackslash\$[a-zA-Z0-9=][a-zA-Z][a-zA-Z0-9=]+'ä
rä`ä\^[a-zA-Z0-9=][a-zA-Z][a-zA-Z0-9=]+'
ä\textbf{Images/links}ä = rä`ä(?:(pic.ä$\mid$ähttpä$\mid$äwwwä$\mid$ä\w+)?\:(//)*)\S+'
ä\textbf{Special char.}ä = rä`ä(\*ä$\mid$ä\[ä$\mid$ä\]ä$\mid$ä=ä$\mid$ä\(ä$\mid$ä\)ä$\mid$ää\textbackslash\$ää$\mid$ä\"ä$\mid$ä\}ä$\mid$ä\{ä$\mid$ä\ä$\mid$ää$\mid$ä\+ä$\mid$ä&ä$\mid$ää\texteuroää$\mid$ää\poundsää$\mid$ä/ä$\mid$ää\textdegreeä)+'

\end{lstlisting}

\subsection{Feature engineering module}
\label{sec:feature_engineering_results}

Side features (1-26 in Table \ref{tab:features}) were obtained with the \texttt{gpt-3.5-turbo}\footnote{Available at \url{https://platform.openai.com/docs/models/gpt-3-5}, February 2024.} \textsc{llm}, selected for its performance and cost-effectiveness (\$0.0005/1,000 tokens for prompt text input and \$0.0015/1,000 tokens for model text output at the time this paper was written).

Listings \ref{lst:prompt_scenario2a} and \ref{lst:response_scenario2a} respectively illustrate the prompt used for feature engineering and the measurements obtained from the \textsc{llm} for features 1-24 in scenario 2 for the particular machine-user utterance pair example at the bottom of Listing \ref{lst:prompt_scenario2a}. The same prompt was used for the whole dialogue (for features 1-24). Listings \ref{lst:prompt_scenario2b} and \ref{lst:response_scenario2b} respectively show the prompt and the results obtained for features 25-26, for the user utterance at the bottom of Listing \ref{lst:prompt_scenario2b}.

Listings \ref{lst:prompt_scenario3} and \ref{lst:response_scenario3} respectively show the prompt for the baseline in scenario 3 (\textit{i}) and an example of the corresponding response by the \textsc{llm}. Finally, Listing \ref{lst:prompt_scenario3_enhanced} shows the prompt designed for the enhanced scenario 3 (\textit{ii}), including the effect of relevant features selected from scenario 2 as described in Section \ref{sec:feature_selection}. The ultimate goal is to improve the ability of the \textsc{llm} to detect cognitive decline directly.

Content-dependent features 27-28 in Table \ref{tab:features} were generated with the \texttt{CountVectorizer} library\footnote{Available at \url{https://scikit-learn.org/stable/modules/generated/sklearn.feature_extraction.text.CountVectorizer.html}, February 2024.} from the \texttt{scikit-learn} Python library. The optimal parameters in Listing \ref{lst:param_char} and \ref{lst:param_word} were calculated using the \texttt{GridSearchCV}\footnote{Available at \url{https://scikit-learn.org/stable/modules/generated/sklearn.model_selection.GridSearchCV.html}, February 2024.\label{fn:gscv}} function from the \texttt{scikit-learn} Python library. A total of \num{1282} features were obtained.

\begin{lstlisting}[frame=single,caption={Prompt used for feature engineering in scenario 2 for features 1-24.}, label={lst:prompt_scenario2a},emphstyle=\textbf,escapechar=ä]
This is a conversation between a bot and a human. Return what
I ask below with a value between 0.0 and 1.0. 
The user has the following problems: impairments, medication
Tell me if the human: has any memory loss, seems disconnected
from the conversation, is tired, hesitates, takes the lead 
in the conversation, has a fluent conversation, seems relaxed,
stressed, happy, exhibits mood swings, the polarity of the 
conversation (negative, neutral, positive), seems sad, 
interacts like an adult, like a child, uses a colloquial 
registry, has comprehension problems, uses an elder registry,
has expression problems, uses a formal registry, repetitive 
language, rush language, short response, bad words, complex 
words.
Respond only in the following JSON format: 
{ä"äamnesia":0.0, ä"disconnection"ä:0.0, ä"fatigue"ä:0.0, 
ä"hesitation"ä:0.0, ä"initiative"ä:0.0, ä"naturalness"ä:0.0, 
ä"relax"ä:0.0, ä"stress"ä:0.0, ä"happiness"ä:0.0, 
ä"mood\_swings"ä:0.0, ä"polarity"ä:0.0,ä"sadness"ä:0.0, 
ä"adult\_registry"ä:0.0, ä"child\_registry"ä:0.0,
ä"colloquial\_registry"ä:0.0, ä"comprehension\_problems"ä:0.0,
ä"elderly\_registry"ä:0.0, ä"expresion\_problems"ä:0.0,
ä"formal\_registry"ä:0.0, ä"repetitive\_language"ä:0.0,
ä"rush\_language"ä:0.0, ä"short\_response"ä:0.0, 
ä"bad\_words"ä:0.0, ä"complex\_words"ä:0.0}.
ALWAYS RETURN A JSON IN THE GIVEN FORMAT WITHOUT ADDING MORE 
TEXT OR MODIFYING THE FIELD NAMES OF THE JSON. DO NOT ANSWER 
ANY QUESTIONS IN THE CONVERSATION.
bot: Did you go anywhere special today Pablo?
human: Yes. Today I went to the beach close to my house with 
my wife, but it was a bit windy.
\end{lstlisting}

\begin{lstlisting}[frame=single,caption={Response received from the transformer model in scenario 2, features 1-24.}, label={lst:response_scenario2a},emphstyle=\textbf,escapechar=ä]
{ä"amnesia"ä:0.0, ä"disconnection"ä:0.0, ä"fatigue"ä:0.2, 
ä"hesitation"ä:0.0, ä"initiative"ä:0.8, ä"naturalness"ä:1.0, 
ä"relax"ä:0.6, ä"stress"ä:0.0, ä"happiness"ä:0.6, 
ä"mood\_swings"ä:0.0, ä"polarity"ä:1.0, ä"sadness"ä:0.3,
ä"adult\_registry"ä:0.7, ä"child\_registry"ä:0.0,
ä"colloquial\_registry"ä:0.0, ä"comprehension\_problems"ä:0.0,
ä"elderly\_registry"ä:0.6, ä"expresion\_problems"ä:0.0,
ä"formal\_registry"ä:0.6, ä"repetitive\_language"ä:0.0,
ä"rush\_language"ä:0.0, ä"short\_response"ä:0.0, 
ä"bad\_words"ä:0.0, ä"complex\_words"ä:0.2}.
\end{lstlisting}

\begin{lstlisting}[frame=single,caption={Prompt used for feature engineering in scenario 2 for features 25-26.}, label={lst:prompt_scenario2b},emphstyle=\textbf,escapechar=ä]
In the following sentence, return the number of adjectives, 
adverbs (negative and positive), nouns and verbs.
Respond only in the following JSON format: 
{ä"adjectives"ä:0.0, ä"adverbs"ä: 0.0, ä"adverbs\_neg"ä:0.0, 
ä"adverbs\_pos"ä:0.0, ä"nouns"ä:0.0, ä"verbs"ä:0.0, 
ä"polar\_response\_yes"ä:0.0, ä"polar\_response\_no"ä:0.0}.
ALWAYS RETURN A JSON IN THE GIVEN FORMAT WITHOUT ADDING MORE 
TEXT OR MODIFYING THE FIELD NAMES OF THE JSON.
Yes. Today I went to the beach close to my house with 
my wife, but it was a bit windy.
\end{lstlisting}

\begin{lstlisting}[frame=single,caption={Response received from the transformer model in scenario 2, features 25-26.}, label={lst:response_scenario2b},emphstyle=\textbf,escapechar=ä]
{ä"adjectives"ä:3.0, ä"adverbs"ä:3.0,ä"adverbs\_neg"ä:0.0, 
ä"adverbs\_pos"ä:0.0, ä"nouns"ä:3.0, ä"verbs"ä:2.0, 
ä"polar\_response\_yes"ä:1.0, ä"polar\_response\_no"ä:0.0}.
\end{lstlisting}

\begin{lstlisting}[frame=single,caption={Prompt used for cognitive detection in scenario 3.}, label={lst:prompt_scenario3},emphstyle=\textbf,escapechar=ä]
This is a conversation between a bot and a human. Return 
what I ask below with a value of 0 or 1.
The user has the
following problems: impairments, medication
Tell me if the human has cognitive impairment.
Respond only in the following JSON format:
{ä"cognitive\_impairment"ä:0}.
ALWAYS RETURN A JSON IN THE GIVEN FORMAT WITHOUT ADDING MORE 
TEXT OR MODIFYING THE FIELD NAMES OF THE JSON. DO NOT ANSWER 
ANY QUESTIONS IN THE CONVERSATION.
bot: Did you go anywhere special today Pablo?
human: Yes. Today I went to the beach close to my house with 
my wife, but it was a bit windy.
...
\end{lstlisting}

\begin{lstlisting}[frame=single,caption={Response received from the transformer model in scenario 3.}, label={lst:response_scenario3},emphstyle=\textbf,escapechar=ä]
{ä"cognitive\_impairment"ä: 0}
\end{lstlisting}

\begin{lstlisting}[frame=single,caption={Prompt used for feature engineering in scenario 3 including most relevant features from scenario 2.}, label={lst:prompt_scenario3_enhanced},emphstyle=\textbf,escapechar=ä]
This is a conversation between a bot and a human. Return
what I ask below with a value of 0 or 1.
The user has the following problems: impairments, medication
Tell me if the human has cognitive impairment based on these 
relevant features extracted:
- mean happiness value.
- maximum short response value.
...
Respond only in the following JSON format:
{ä"cognitive\_impairment"ä:0}.
ALWAYS RETURN A JSON IN THE GIVEN FORMAT WITHOUT ADDING MORE 
TEXT OR MODIFYING THE FIELD NAMES OF THE JSON. DO NOT ANSWER 
ANYQUESTIONS IN THE CONVERSATION.
bot: Did you go anywhere special today Pablo?
human: Yes. Today I went to the beach close to my house with 
my wife, but it was a bit windy.
\end{lstlisting}
\begin{lstlisting}[frame=single,caption={Parameter selection for char-grams (best values in bold).}, label={lst:param_char},emphstyle=\textbf,escapechar=ä]
max_df = [0.25, ä\bf{0.5}ä, 0.75],
min_df = [ä\bf{0.05}ä, 0.075, 0.1],
ngram_range = [ä\bf{(3,4)}ä, (3,6), (4,6)],
max_features = [500, ä\bf{5000}ä, None ]
\end{lstlisting}

\begin{lstlisting}[frame=single,caption={Parameter selection for word-grams (best values in bold).}, label={lst:param_word},emphstyle=\textbf,escapechar=ä]
max_df = [ä\bf{0.25}ä, 0.5, 0.75],
min_df = [ä\bf{0.05}ä, 0.075, 0.1],
ngram_range = [(1,1), ä\bf{(1,3)}ä, (2,4)],
max_features = [ä\bf{500}ä, 5000, None],
\end{lstlisting}

\subsection{Feature analysis \& selection module}
\label{sec: feature_selection_results}

In scenarios 1 and 2, features were selected with the \texttt{SelectFromModel}\footnote{Available at \url{https://scikitlearn.org/stable/modules/generated/sklearn.feature_selection.SelectFromModel.html}, February 2024.} transformer wrapper using \textsc{rf}, with default parameter settings. The result of the selection was:

\begin{itemize}
 \item \textbf{Scenario 1}: \num{435} char-gram and word-gram features.
 \item \textbf{Scenario 2}: \num{262} feature components comparing statistics from the current session with those of other sessions plus \num{7} feature components corresponding to statistics from the current session.
\end{itemize}

The Pearson correlation coefficient\footnote{Available at \url{https://scikit-learn.org/stable/modules/generated/sklearn.feature_selection.r_regression.html}, February 2024.} was employed in scenario 3 to further select features from scenario 2 with a stronger direct or indirect relationship with the target. Only features with a correlation over 0.45 were selected based on empirical tests (see Table \ref{tab:pearson_correlation_results}). All the features corresponded to feature components, comparing statistics from the current session and those from other sessions.

\begin{table}[!htpb]
\centering
\caption{\label{tab:pearson_correlation_results}Pearson correlation results.}
\begin{tabular}{llc}
\toprule
\textbf{Feature} & \begin{tabular}[c]{@{}p{3.5cm}@{}} \centering\textbf{Session comparison statistic}\end{tabular} & \begin{tabular}[c]{@{}p{3.5cm}@{}} \centering\textbf{Correlation with target}\end{tabular} \\
\midrule
9 (maximum) & Mean & -0.46 \\
22 (maximum) & Mean & 0.49 \\
26 (nouns minimum) & Mean & -0.47 \\
26 (nouns mean) & Mean & -0.48 \\
26 (nouns percentile 25) & Mean & -0.47 \\
26 (nouns percentile 50) & Mean & -0.48 \\
26 (nouns mean) & Percentile 25 & -0.49 \\
26 (nouns percentile 25) & Percentile 25 & -0.46 \\
22 (maximum) & Percentile 50 & 0.48 \\
22 (mean) & Percentile 75 & 0.46 \\
22 (maximum) & Percentile 75 & 0.47\\
26 (nouns minimum) & Percentile 75 & -0.46 \\
20 (maximum) & Maximum & 0.47 \\
22 (percentile 25) & Maximum & 0.46 \\
26 (nouns percentile 50) & Minimum & -0.48 \\
\bottomrule
\end{tabular}
\end{table}

\subsection{Classification module}
\label{sec:evaluation_results}

The implementations of the \textsc{ml} algorithms were: 

\begin{itemize}
 \item \textsc{nb}. Gaussian Naive Bayes\footnote{Available at \url{https://scikit-learn.org/stable/modules/generated/sklearn.naive_bayes.GaussianNB.html}, February 2024.} from the \texttt{scikit-learn} Python library.

 \item \textsc{dt}. \texttt{DecisionTreeClassifier} \footnote{Available at \url{https://scikit-learn.org/stable/modules/generated/sklearn.tree.DecisionTreeClassifier.html}, February 2024.} from the \texttt{scikit-learn} Python library.

 \item \textsc{rf}. \texttt{RandomForestClassifier} \footnote{Available at \url{https://scikit-learn.org/stable/modules/generated/sklearn.ensemble.RandomForestClassifier.html}, February 2024.} from the \texttt{scikit-learn} Python library.
\end{itemize}

For each implementation, optimal hyperparameters were tuned with the aforementioned \texttt{GridSearchCV} method using $10$-fold cross-validation. Listings \ref{lst:hyperparam_dt}-\ref{lst:hyperparam_rf} show the ranges of values explored for scenarios 1 and 2. The final parameters selected were the following:

\begin{description}
 \item \textbf{Scenario 1}
 \begin{itemize}
 \item \textbf{DT}: splitter=random, class\_weight=None, max\_features=log2, max\_depth=100, min\_samples\_split=\num{0.1}, min\_samples\_leaf=\num{0.001}

 \item \textbf{NB}: var\_smoothing=1e-9

 \item \textbf{RF}: n\_estimators=\num{75}, class\_weight=None, max\_features=log2, max\_depth=\num{100}, min\_samples\_split=\num{5}, min\_samples\_leaf=\num{1}
 \end{itemize}

 \item \textbf{Scenario 2}
 \begin{itemize}
 \item \textbf{DT}: splitter=best, class\_weight=None, max\_features=None, max\_depth=100, min\_samples \_split=\num{0.001}, min\_samples\_leaf=\num{0.001}

 \item \textbf{NB}: var\_smoothing=1e-6

 \item \textbf{RF}: n\_estimators=\num{50}, class\_weight=balanced, max\_features=log2, max\_depth=\num{10}, min\_samples\_split=\num{2}, min\_samples\_leaf=\num{1}
 \end{itemize}
\end{description}

\begin{lstlisting}[frame=single,caption={Hyperparameter selection for \textsc{dt}.}, label={lst:hyperparam_dt},emphstyle=\textbf,escapechar=ä]
splitter = [best, random],
class_weight = [None, balanced],
max_features = [None, sqrt, log2],
max_depth = [1, 100, None],
min_samples_split = [0.001, 0.1, 1],
min_samples_leaf = [0.001, 0.1, 1]
\end{lstlisting}

\begin{lstlisting}[frame=single,caption={Hyperparameter selection for \textsc{nb}.}, label={lst:hyperparam_nb},emphstyle=\textbf,escapechar=ä]
var_smoothing = [1e-9, 1e-8, 1e-7, 1e-6, 1e-5, 1e-4, 1e-3,
1e-2, 1e-1]
\end{lstlisting}

\begin{lstlisting}[frame=single,caption={Hyperparameter selection for \textsc{rf}.}, label={lst:hyperparam_rf},emphstyle=\textbf,escapechar=ä]
n_estimators = [50, 75, 100],
class_weight = [balanced, None],
max_features = [sqrt, log2, None],
max_depth = [5, 10, 100, None],
min_samples_split = [2, 5, 10],
min_samples_leaf = [1, 2, 5],
\end{lstlisting}

Tables \ref{tab:classification_results_1} and \ref{tab:classification_results_2} respectively present the results of the performance evaluation for scenarios 1 and 2 with $10$-fold cross-validation, implemented with the \texttt{scikit-learn} Python library\footnote{Available at \url{https://scikit-learn.org/stable/modules/generated/sklearn.model_selection.cross_val_predict.html}, February 2024.}. This evaluation method divides the data set into 10 folds (9 for training and 1 for testing). This split is repeated 10 times using different partitions without overlapping testing sets to minimize evaluation bias. The final results are an average of results for the different splits. Our experiments used 471 and 52 samples to train and test the models in each split.

\begin{table}[!htbp]
\centering
\caption{\label{tab:classification_results_1}Classification results for scenario 1 (Yes: cognitive impairment, No: otherwise).}
\begin{tabular}{ccccccccS[table-format=3.2]}
\toprule
\bf Model & \bf Acc. & \multicolumn{3}{c}{\bf Precision} & \multicolumn{3}{c}{\bf Recall} & {\bf Time (s)}\\
\cmidrule(lr){3-5}
\cmidrule(lr){6-8}
 & & Macro & No & Yes & Macro & No & Yes \\
\midrule
\textsc{rf} & \textbf{76.67} & \textbf{77.18} & 75.58 & \textbf{78.77} & 74.96 & \textbf{87.24} & 62.67 & 1.01 \\
\textsc{dt} & 65.39 & 64.70 & 66.76 & 62.64 & 63.32 & 78.19 & 48.44 & 0.44 \\
\textsc{nb} & 74.95 & 74.83 &\textbf{81.27} & 68.39 & \textbf{75.30} & 72.82 & \textbf{77.78} & 0.40 \\
\bottomrule
\end{tabular}
\end{table}

\begin{table}[!htbp]
\centering
\caption{\label{tab:classification_results_2}Classification results for scenario 2 (Yes: cognitive impairment, No: otherwise).}
\begin{tabular}{ccccccccS[table-format=3.2]}
\toprule
\bf Model & \bf Acc. & \multicolumn{3}{c}{\bf Precision} & \multicolumn{3}{c}{\bf Recall} & {\bf Time (s)}\\
\cmidrule(lr){3-5}
\cmidrule(lr){6-8}
 & & Macro & No & Yes & Macro & No & Yes \\
\midrule
\textsc{rf} & \textbf{98.47} & \textbf{98.49} & \textbf{98.33} & \textbf{98.65} & \textbf{98.39} & \textbf{98.99} & \textbf{97.78} & 0.54 \\
\textsc{dt} & 93.50 & 93.46 & 93.71 & 93.21 & 93.27 & 94.97 & 91.56 & 0.25 \\
\textsc{nb} & 90.63 & 90.34 & 94.31 & 86.36 & 90.91 & 88.93 & 92.89 & 0.18 \\
\bottomrule
\end{tabular}
\end{table}

In scenarios 1 and 2, the best accuracy levels were achieved by the \textsc{rf} model, 
\SI{76.67}{\percent} and \SI{98.47}{\percent}, respectively. In all cases, scenario 2 outperformed scenario 1. This is interesting for free dialogues from a practical perspective since scenario 2 is context-independent. 

The results for scenario 3 were significantly worse than those obtained in scenario 2. The baseline approach based only on Chat\textsc{gpt} directly applied to free dialogues only attained \SI{57.17}{\percent} accuracy. However, when knowledge about context-independent high-level reasoning features was transferred to the \textsc{llm} by prompt engineering, the accuracy rate rose to \SI{61.19}{\percent}. We concluded that pre-trained models cannot replace specialized models but are extremely useful for producing valuable context-independent features. 

In terms of related solutions in the literature, our proposal was more accurate (\SI{26.19}{\percent}) than the approach described by \citet{Agbavor2022}. Moreover, it was associated with respective improvements of +\SI{16.33}{\percent} and +\SI{13.65}{\percent} for the detection of absent and existing cognitive impairment compared to the solution described by \citet{Qiao2021}. On comparing our findings with those of \citet{Yuan2020}, our proposal was \SI{3.19}{\percent} and \SI{14.48}{\percent} more accurate for detecting absent and existing cognitive impairment classes, respectively. The corresponding improvements over the proposal of \citet{zhu2021wavbert}, were +\SI{10.01}{\percent} and +\SI{20.64}{\percent}. Finally, our proposal achieved better results than \citet{wang2023text} (+\SI{6.67}{\percent} accuracy), the most closely related work, when different types of content (chatbot-human dialogues and doctor-patient dialogues) were analyzed. Nevertheless, when Chat\textsc{gpt} rather than \textsc{hlr} features derived from doctor-patient dialogues were used as the content source, the improvement grew to +\SI{37.28}{\percent}.

The following paragraph describes how our system's mean token consumption was calculated. In scenario 2 of our analysis, we used 1400 and 498 characters for the prompts in listings \ref{lst:prompt_scenario2a} and \ref{lst:prompt_scenario2b}, respectively. On average, the complete dialogue by the conversational assistant with a user comprised 15.72 utterances with 909 characters. Also, on average, each artificially generated utterance had 57.82 characters. The generated outputs approximately consumed 483 characters for the first prompt (see Listing \ref{lst:response_scenario2a}) and 147 characters for the second (see Listing \ref{lst:response_scenario2a}). Therefore, the average total numbers of input and output characters were 33,325.19 (1,400 + 909 + (1,400 + 57.82) $\times$ 15.72 + (498 + 17.22) $\times$ 15.72) and 10,386.60 (483 $\times$ 16.72 + 147 $\times$ 15.72), respectively. Since the character/token ratio in Open \textsc{ai}\footnote{Available at \url{https://platform.openai.com/tokenizer}, February 2024.} is approximately 4/1, the cost per user session was moderate (\textit{i.e.}, \$0.0081). It did not compromise the scalability of the study nor its possible transfer to industry.

\subsection{Explainability module}
\label{sec:explainability_results}

To explain the model's predictions, we used the template in Listing \ref{lst:exp_template_en}. This template details the most relevant \textless $features$\textgreater\ used to classify a \textless $user$\textgreater\ from a given \textless $conversation$\textgreater\ (a dialogue session in our case). The features used in this template are the six relevant features selected using the meta-transformer wrapper, comprising several components described in Section \ref{sec:feature_engineering}. Only the components with the highest and lowest values were considered for explainability after discarding the minimum and maximum statistics. The values of these components were sorted, and the three largest and three most minor were used.

\begin{lstlisting}[frame=single,caption={Natural language explanation template (in English).}, label={lst:exp_template_en},emphstyle=\textbf,escapechar=ä]
The classification of the conversation

ä\textcolor{blue}{\textless $conversation$\textgreater}ä 

as ä\textcolor{blue}{\textless $cognitive impairment$\textgreater}ä can be explained by 
the presence of these features ä\textcolor{blue}{\textless $features$\textgreater}ä and a lack of 
ä\textcolor{blue}{\textless $features$\textgreater}ä.

\end{lstlisting}

\begin{figure}
\centering
\begin{subfigure}{.5\textwidth}
 \centering
 \includegraphics[width=.7\linewidth]{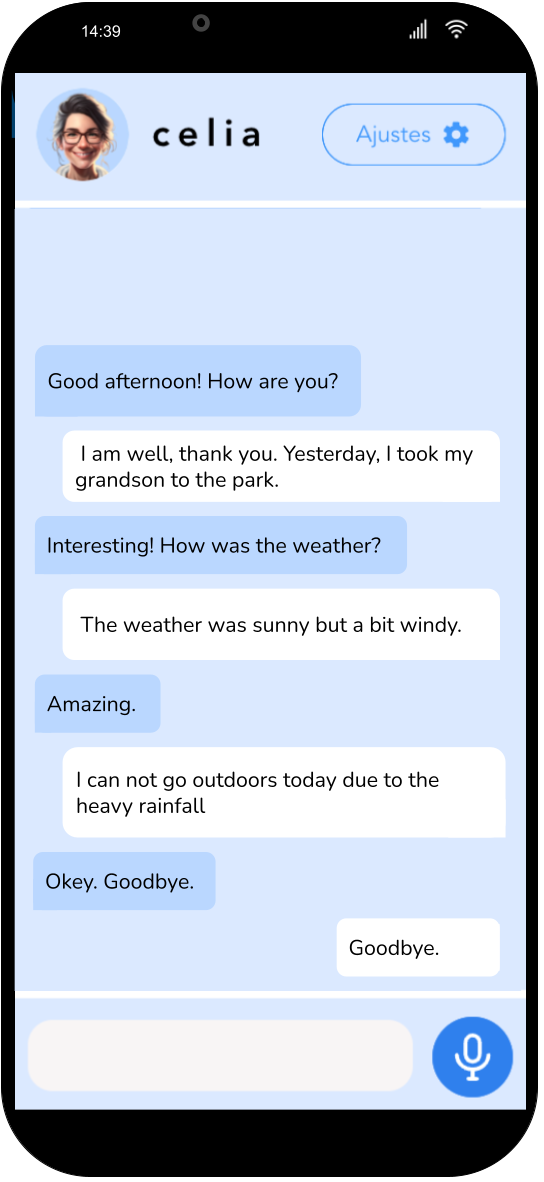}
 \caption{Conversation.}
 \label{fig:explainable1}
\end{subfigure}%
\begin{subfigure}{.5\textwidth}
 \centering
 \includegraphics[width=.7\linewidth]{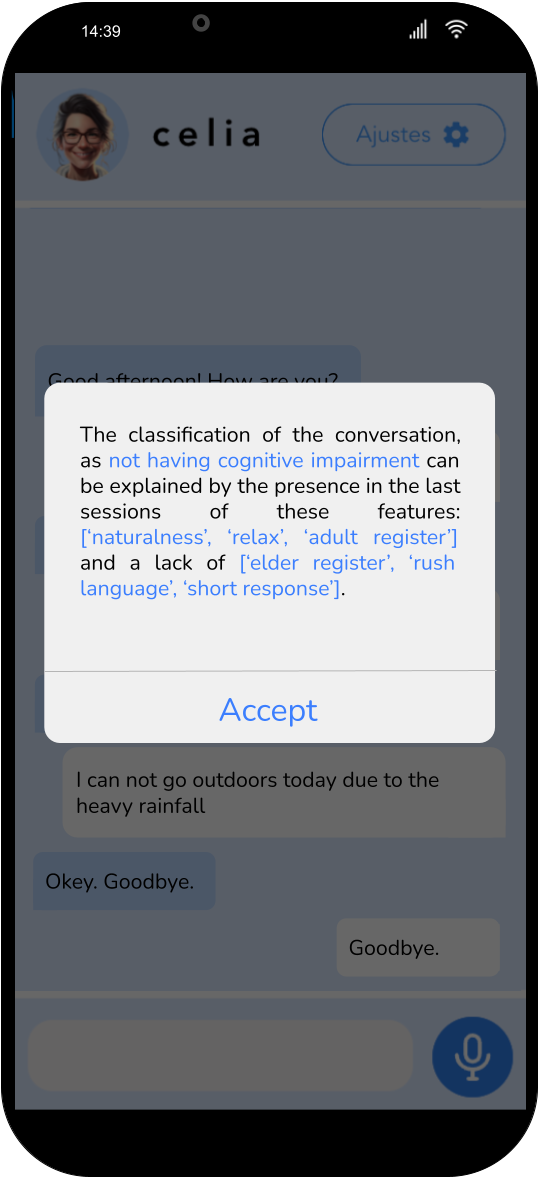}
 \caption{Explainability.}
 \label{fig:explainable2}
\end{subfigure}
\caption{Celia application with explainability capabilities.}
\label{fig:explainable}
\end{figure}

Listing \ref{lst:res_template} and Figure \ref{fig:explainable} show a real-life example of a generated explanation for a classified conversation. In this case, the system did not detect cognitive impairment because, among other reasons, the user was relaxed and natural, used an adult register, and did not respond concisely (recall that the explainability template only uses the most relevant features to classify a particular individual). This explanation in natural language is descriptive and understandable, even without expert knowledge in healthcare or \textsc{ml}.

\begin{lstlisting}[frame=single,caption={Explanation example.}, label={lst:res_template},emphstyle=\textbf,escapechar=ä]
The classification of the conversation 

ä\textcolor{blue}{bot: Good afternoon! How are you?}ä

ä\textcolor{blue}{human: I am well, thank you. Yesterday, I took my grandson to the park.}ä

ä\textcolor{blue}{bot: Interesting! How was the weather?}ä

ä\textcolor{blue}{human: The weather was sunny but a bit windy.}ä

ä\textcolor{blue}{bot: Amazing.}ä

ä\textcolor{blue}{human: I cannot go outdoors today due to the heavy rainfall.}ä

ä\textcolor{blue}{bot: Okey. Goodbye.}ä

ä\textcolor{blue}{human: Goodbye.}ä

as ä\textcolor{blue}{not having cognitive impairment}ä can be explained by 
the presence in the last sessions of these features: 
ä\textcolor{blue}{[`naturalness', `relax', `adult register']}ä 
and a lack of ä\textcolor{blue}{[`elder register', `rush language', `short response']}ä.
\end{lstlisting}

\section{Conclusions}
\label{sec:conclusions}

This work is the first to apply \textsc{ml} techniques to high-level reasoning features extracted using \textsc{llm}s from free dialogues to detect cognitive decline with the Celia entertainment chatbot. High-level reasoning features are context-independent and are, therefore, more widely applicable for characterizing free dialogues than word-embedding or content-dependent features. They also support significantly more interpretable descriptions of the decisions using explainability techniques. The essential advantage of free dialogues with engaging systems is that they may encourage end users to participate in longitudinal studies, which currently need to be more feasible in healthcare systems due to the high cost of manual screening.

A mixed approach combining a specialized \textsc{ml} with feature extraction using the Chat\textsc{gpt}
\textsc{llm} outperformed direct evaluation of cognitive decline by Chat\textsc{gpt}, even when prompt engineering was used to boost Chat\textsc{gpt} with the best features of the \textsc{ml} model. The performance levels achieved are remarkable. 

In future work, we plan to train our models using a streaming mode that incorporates the session history. We will also employ new methodologies (\textit{e.g.}, reinforcement learning) to study different motivational topics and analyze how new features influence prediction outcomes. Note that we are only using a  \textsc{llm} to obtain precise answers on the semantic relationships between different pieces of text. Thus, our approach is less affected by certain issues of generative \textsc{llm}s  (\textit{e.g.}, context and memory management, layer pruning, hallucination issues, etc.) than the direct application of a  \textsc{llm} in scenario 3. However, in future work, we plan to analyze the sensitivity of these issues.

\section{Declarations}

\subsection*{Funding}

This work was partially supported by (\textit{i}) Xunta de Galicia grants ED481B-2022-093, ED481D 2024/014, and ED431C 2022/04, Spain; (\textit{ii}) Ministerio de Ciencia e Innovación grant TED2021-130824B-C21, Spain; and (\textit{iii}) University of Vigo/CISUG for open access charge.

\subsection*{Competing interests}

The authors have no competing interests to declare relevant to this article's content.

\subsection*{Authors contribution}

\textbf{Francisco de Arriba-Pérez}: Conceptualization, Methodology, Software, Validation, Formal analysis, Investigation, Resources, Data Curation, Writing - Original Draft, Writing - Review \& Editing, Visualization, Supervision, Project administration, Funding acquisition. \textbf{Silvia García-Méndez}: Conceptualization, Methodology, Software, Validation, Formal analysis, Investigation, Resources, Data Curation, Writing - Original Draft, Writing - Review \& Editing, Visualization, Supervision, Project administration, Funding acquisition. \textbf{Javier Otero-Mosquera}: Software, Data Curation, Writing - Review \& Editing. \textbf{Francisco J. González-Castaño}: Conceptualization, Methodology, Writing - Review \& Editing, Supervision, Funding acquisition.

\subsection*{Data availability}

Data will be made available on reasonable request.

\bibliography{3_bibliography}

\end{document}